\documentclass[letterpaper]{article} 
\usepackage{aaai24}  
\usepackage{times}  
\usepackage{helvet}  
\usepackage{courier}  
\usepackage[hyphens]{url}  
\usepackage{graphicx} 
\usepackage{natbib}  
\usepackage{caption} 
\usepackage{algorithm}
\usepackage{algorithmic}
\usepackage{xcolor}
\usepackage{multirow}
\usepackage{makecell}
\usepackage[most]{tcolorbox}
\usepackage{url}
\usepackage{tabularx}
\usepackage{diagbox}
\usepackage{booktabs} 
\definecolor{mygreen}{RGB}{209,255,200}
\definecolor{myred}{RGB}{255,205,196}

\definecolor{Lavender}{RGB}{230, 230, 250}
\definecolor{YellowOrange}{RGB}{255, 204, 0}

\colorlet{LightLavender}{Lavender!30!}
\colorlet{LightRed}{YellowOrange!20!}
\colorlet{LightOrange}{myred!20!}
\tcbset{on line, 
    boxsep=1.0pt, left=0pt, right=0pt,top=0pt,bottom=0pt,
    colframe=white, colback=LightRed,  
    highlight math style={enhanced}
}
\usepackage{newfloat}
\usepackage{listings}
\DeclareCaptionStyle{ruled}{labelfont=normalfont,labelsep=colon,strut=off} 
\floatstyle{ruled}
\newfloat{listing}{tb}{lst}{}
\floatname{listing}{Listing}
\title{Watermarking Conditional Text Generation for AI Detection: \\
Unveiling Challenges and a Semantic-Aware Watermark Remedy}
\author{
    Yu Fu\textsuperscript{\rm 1},
    Deyi Xiong\textsuperscript{\rm 2},
    Yue Dong\textsuperscript{\rm 1}\thanks{Corresponding Author}
}
\affiliations{
    \textsuperscript{\rm 1}University of California Riverside\\
    \textsuperscript{\rm 2}Tianjin University \\

    yfu093@ucr.edu\textsuperscript{\rm 1},
    dyxiong@tju.edu.cn\textsuperscript{\rm 2},
    yue.dong@ucr.edu\textsuperscript{\rm 1}
}

\usepackage{bibentry}

\begin{document}

\maketitle
\pagestyle{plain} 
\begin{abstract}

To mitigate potential risks associated with language models (LMs), recent AI detection research proposes incorporating watermarks into machine-generated text through random vocabulary restrictions and utilizing this information for detection. In this paper, we show that watermarking algorithms designed for LMs cannot be seamlessly applied to conditional text generation (CTG) tasks without a notable decline in downstream task performance. To address this issue, we introduce a simple yet effective semantic-aware watermarking algorithm that considers the characteristics of conditional text generation with the input context. Compared to the baseline watermarks, our proposed watermark yields significant improvements in both automatic and human evaluations across various text generation models, including BART and Flan-T5, for CTG tasks such as summarization and data-to-text generation. Meanwhile, it maintains detection ability with higher $z$-scores but lower AUC scores, suggesting the presence of a detection paradox that poses additional challenges for watermarking CTG.\footnote{https://github.com/FYYFU/semantic-watermark}

\end{abstract}

\begin{figure}[t]
    \centering
    \includegraphics[scale=0.8]{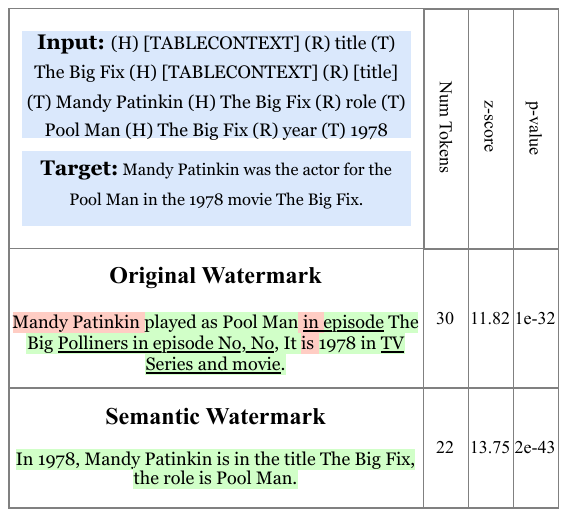}
    \caption{The outputs with the original watermark (OW) \citep{kirchenbauer2023watermark} and our proposed semantic-aware watermark (SW) on a test example from DART -- a data-to-text generation benchmark -- with parameters $\gamma=0.1$ and $\delta=5$. We expect $\sim$ 90\% of human-generated texts from the red list, whereas AI primarily utilizes the green list. Both watermarks yield high $z$-scores ($z>4$), indicating strong watermark strength for detection. Yet, OW forces the algorithm to generate from the red list due to randomly assigning key source entities (Mandy Patinkin) to it. As $\delta$ increases (towards a hard watermark), excluding these red tokens risks more hallucinations (words with underline).}
    \label{fig:enter-label}
\end{figure}

\section{Introduction}

Language Models (LMs) have demonstrated remarkable effectiveness in generating content that closely resembles human performances across diverse tasks \cite{tan2023chatgpt, dong2023selfcollaboration, liu2023code}. As large-scale models such as ChatGPT \cite{openai2021chatgpt} evolve and produce increasingly human-like content, concerns have surged around potential limitations and risks tied to their use \cite{bender2021dangers},  including hallucination \citep{alkaissi2023artificial}, bias and toxicity \citep{deshpande2023toxicity}, failure in commonsense reasoning \cite{bian2023chatgpt}, and misinformation and malicious use \citep{openai2023gpt4}.

To mitigate potential risks associated with LMs, it's crucial to develop methods that differentiate between AI and human-generated content.  Current AI-detection tools primarily rely on perplexity-based classifiers, assuming lower perplexity in AI-generated text \cite{solaiman2019release,jawahar-etal-2020-automatic,mitchell2023detectgpt,mitrović2023chatgpt}. Conversely, an alternative approach is to inject watermarks during generation for subsequent detection. For instance, \citet{kirchenbauer2023watermark} proposed using hash function to randomly bifurcate the vocabulary into ``green'' and ``red'' lists at each decoding step, serving as watermarks. This watermark provides reliable detection signals without the need to train a classifier and produces high-quality generated texts with only a slight perplexity drop in language modeling.

Different from existing research, our focus is on watermarks for conditional text generation (CTG), and we unveil the challenges associated with the use of watermarks \citep{kirchenbauer2023watermark}. Our research findings suggest that \textit{LM watermarking algorithms cannot be seamlessly applied to CTG tasks without a notable decline in performance}: the omission of task-specific considerations leads to significant decreases observed -- up to 96.99\% drop with hard watermarks and 27.54\% drop with soft watermarks  -- in conditional generation tasks including summarization \citep{see-etal-2017-get, narayan-etal-2018-dont} and data-to-text generation \citep{gardent-etal-2017-webnlg,nan-etal-2021-dart}. Figure \ref{fig:enter-label} illustrates an example where the randomly bifurcated red list \citep{kirchenbauer2023watermark} contains key entities from the source that has to be generated for the data-to-text generation task; the mismatch between context and watermark not only impairs detection but also introduces 12 hallucinated words in a 30-token generation.  

To enhance the effectiveness of watermarks for CTG, we propose a simple yet effective semantic-aware watermarking algorithm that leverages hash function to embed watermarks, while also taking into account the input context and the distinctive characteristics of conditional generation tasks. In particular, we strategically bifurcate the vocabulary to balance randomness and semantic relatedness to the input source using word vector similarity. These semantically-related tokens can efficiently cover a substantial portion of the information that needs to be generated in conditional text generation tasks. Consequently, their inclusion in the ``green list" acts as a buffer, reducing the adverse impact of adding watermarks. Compared to the baseline watermarks, our watermark maintains detection ability with higher $z$-scores but lower AUC scores (higher values are better for both). This suggests the presence of a detection paradox that introduces additional challenges for watermarking CTG: \textit{the prevalent human habit of using tokens identical/similar to the input for CTG complicates the detection of watermarks and implies a trade-off between detection ability and metric score}.

Our contributions can be summarized as follows:

\begin{itemize}
    \item We show that directly applying \citet{kirchenbauer2023watermark}'s watermarking method to conditional text generation tasks, without task-specific considerations, can lead to a significant performance drop (up to 96.99\%). This significant decline is observed across multiple tasks like summarization and data-to-text generation, and various text generation models such as BART and Flan-T5.
    \item We propose a semantic-aware watermarking algorithm that utilizes hash function while considering the input context of CTG tasks. Automatic and human evaluations on multiple datasets and models indicate that our method effectively mitigates quality degradation associated with the use of watermarks, while minimizing the trade-off in detection. 
\end{itemize}

\section{Related Work}
\paragraph{Automatic Detection}
The detection of AI-generated text, particularly in the context of large language models (LLMs), has recently attracted significant research interest  \cite{bakhtin2019real, schuster-etal-2020-limitations, frohling2021feature, sadasivan2023aigenerated, mitchell2023detectgpt}. Previous approaches have primarily focused on leveraging the perplexities of generated texts for detection. For example, \citet{solaiman2019release} utilized a classifier to evaluate the total log probability of the text, using it as a means to determine whether the content originated from a machine. Building on this premise, \citet{mitchell2023detectgpt} further validated that the log probability of machine-generated text diminishes upon perturbation, while the log probability of human-written text remains unpredictable when perturbed.

\paragraph{Watermarking} 
There has been a recent emergence of watermarking specific patterns into language models for AI detection.  \citet{zhao2023protecting} focused on injecting secret sinusoidal signals into the decoding steps for each target token by modifying the corresponding probability distribution.  \citet{kirchenbauer2023watermark} proposed a method that randomly bifurcates the vocabulary and modifies the probability distribution during each decoding step, thereby ensuring the inclusion of detectable patterns (watermarks) in the generated text.  Subsequent work by \citet{lee2023wrote} optimized this watermark based on entropy, while \citet{wang2023codable} introduced a novel watermarking scheme that enables the watermark to convey meaningful messages such as user IDs or LLM names, expanding its purpose beyond merely indicating machine-generated text.  On the other hand, \citet{yoo-etal-2023-robust} and \citet{yang2023watermarking} focused on incorporating watermarks through post-processing, allowing for watermarking even in the context of black-box LLMs.  
In contrast to the aforementioned papers, our focus is on watermarking for conditional text generation (CTG) tasks, specifically discussing challenges in applying watermarks designed for LLMs to CTG tasks, and proposing watermarks that incorporate task-specific characteristics that account for input context for CTG.

\section{Method}

This section provides an overview of the basic principles of watermarks, elaborates on our proposed semantic-aware method, and discusses how it's integrated into the watermarking procedure for CTG.

\paragraph{Original Watermark}  Considering a language model with parameters denoted by $\theta$, the probability distribution for the $t$-th token in  sequence $\mathbf{S}=\{s_1, s_2, \dots, s_{|\mathbf{S}|}\}$ can be formulated as :

\begin{equation}
    p(s_{t}) = p_{\theta}(s_t | s_{<t})
    \label{lms}
\end{equation}
By considering all preceding tokens, language models (LMs) generate a probability distribution across the vocabulary and sample tokens accordingly.

Watermarking is a technique designed to incorporate robust detection signals into machine-generated text. \citet{kirchenbauer2023watermark} propose two methods, namely hard and soft watermarks, for adding watermarks to text by imposing vocabulary restrictions during each decoding step. Specifically, the ``Hard Red List" watermarking algorithm randomly divides the vocabulary into ``green" and ``red" lists using a hash function and previously generated tokens. During the generation process, only tokens from the green list can be selected for the $t$-th position. To detect the presence of the watermark in the generated text, a statistical analysis such as the \emph{one proportion z-test} can be employed.

However, randomly partitioning the vocabulary and solely selecting words from the green list can hinder the generation of crucial tokens that are not included in the green list. As an alternative, the ``Soft Red List" watermarking approach introduces a constant $\delta$ to the logit $l^{(t)}_k$ of tokens in the green list during prediction:
\begin{equation}
    p^{(t)}_k = \exp (l^{(t)}_k + \delta) / \sum_{i} \exp (l^{(t)}_i)
\end{equation}
This adjustment ensures that even if there are deterministic tokens not included in the green list, they can still be generated. We observe that hard watermarks can be seen as a special case of soft watermarks, achieved by adding a large $\delta$ to the tokens in the green list. Therefore, we choose soft watermarking algorithm as the unified formulation in our paper.

\subsection{Semantic-Aware Watermark}
In contrast to text generation tasks involving language models, conditional text generation (CTG) tasks often exhibit significant textual overlap, either at the token level or the semantic level.  For instance, \citet{chen-etal-2020-cdevalsumm} demonstrate that in the CNN/DailyMail dataset \citep{see-etal-2017-get}, over 80\% of the tokens found in the summary can be located within the original document. Even in the case of the XSUM dataset \citep{narayan-etal-2018-dont}, known for its ``abstractive" nature, this percentage remains above 60\%. Consequently, random watermarking algorithms, which bifurcate the vocabulary arbitrarily at each decoding step, can drastically impair the performance of generation models.

Considering this characteristic of CTG tasks, we propose a simple yet effective semantic-aware watermarking method to enhance performance. Our approach uses the input context to extract semantically related tokens, measured by word vector similarity to the source.  By incorporating semantically related tokens as a constraint, we ensure the quality of the generated output. We then apply the original watermark and randomly bifurcate the remaining vocabulary.

\begin{algorithm}[t]
\begin{small}
\caption{Semantic-Aware Watermark}
\textbf{Input}: Input sequence $\mathbf{x} = \{x_1, x_2, \dots, x_{|\mathbf{x}|} \}$ \\ \textbf{Parameter}: Conditional model $p_{\theta}$, green list size: $\gamma \in (0,1)$, hardness parameter: $\delta > 0$ cluster parameter: $k \in [1, 2, 5, 10]$ \\
\textbf{Output}: Watermarked text $y$\\

\begin{algorithmic}[1]
    \STATE \emph{Get word embeddings and compute the $|V| \times |V|$ word similarity matrix $\mathbf{M}$}.
    
    \STATE \emph{Using input sequence $\mathbf{x}$ and parameter $k$ to get semantically related tokens $S$ and insert them to ``green list" $G$}.
    
    \FOR{$t = 0,1, \cdots $}
    \STATE Apply the conditional model to input sequence $\mathbf{x}$ and get a logit vector $l^{(t)}$ over the vocabulary $V$.
    \STATE Compute a hash of token $y_{t-1}$ and use it to seed a random number generator.
    \STATE \emph{Using the random number generator and partition the remaining vocabulary into $G$ of size \underline{$\gamma |V| - len(S)$} and a ``red list" $R$ of size \underline{$(1-\gamma) |V|$}}.
    \STATE Add $\delta$ to each green list logit. Apply these modified logits to get a probability distribution over V.
     
 \STATE
 \hspace{-1mm}
    $$\hat p^{(t)}_k =
     \bigg \{ \begin{array}{ll}
         \frac{ \exp(l^{(t)}_k+\delta)}{\sum_{i\in R} \exp(l^{(t)}_i)+\sum_{i\in G} \exp(l^{(t)}_i+\delta)}, &  k\in G \\
         \\
        \frac{ \exp(l^{(t)}_k)}{\sum_{i\in R} \exp(l^{(t)}_i)+\sum_{i\in G} \exp(l^{(t)}_i+\delta)}, & k\in R
    \end{array} 
    $$

\STATE Sample the next token $y_{t}$ according to watermarked distribution $\hat{p}^{(t)}$.
    \ENDFOR
    
\end{algorithmic}
\label{main_algorithm}
\end{small}
\end{algorithm}

To implement this approach, we tokenize the input sequence $\mathbf{x}$ to $\hat{\mathbf{x}} = \{\hat{x}_1, \hat{x}_2, \dots, \hat{x}_{|\mathbf{\hat{x}}|} \}$. Next, the tokenized sequence $\hat{\mathbf{x}}$ is transformed into contextualized vector representations using the model's embedding layer. Integrating input information into the watermark's green list is a direct and crucial step (step 2 in Algorithm 1), consistent with the requirements of CTG tasks where the output is dependent on the input. However, it's crucial to note that output information isn't solely determined by the input. Thus, relying exclusively on input as a constraint may not yield optimal results. To overcome this limitation, we broaden the constraints by incorporating token embeddings to measure token similarities.

\begin{table*}[ht]
\centering
\resizebox{0.97\textwidth}{!}{
\begin{tabular}{lllccc|llll}
\toprule
Dataset  & Model   & Method   & R-1 & R-2 & R-L & Dataset    & Model  & Method  & BLEU  \\ 
\midrule
\multirow{12}{*}{\textbf{CNN}}  & \multirow{6}{*}{BART-large}  & NW & 43.80 & 20.88 & 40.73  & \multirow{12}{*}{\textbf{DART}} & \multirow{6}{*}{BART-large} & NW  & 47.78 \\
&  & OW (Hard) & 33.38  & 8.73 & 30.61 &  &  & OW (Hard) & 6.65 \tcbhighmath[colback=LightOrange]{\downarrow 86.1\%} \\
&  & SW (Hard) & \textbf{43.46} &\textbf{20.75} & \textbf{40.45} &   &   & SW (Hard) & \textbf{41.04} \tcbhighmath[colback=LightOrange]{\downarrow 14.1\%}  \\
&  & OW (Soft) & 42.46 & 18.33 & 39.52 &   &  & OW (Soft)& 37.06 \tcbhighmath[colback=LightOrange]{\downarrow 22.4\%}\\
&  & SW (Soft) & \textbf{43.50} & \textbf{20.83} & \textbf{40.62} &     &     & SW (Soft) & \textbf{44.04} \tcbhighmath[colback=LightOrange]{\downarrow 7.8\%} \\ \cmidrule{2-6} \cmidrule{8-10} 
& \multirow{6}{*}{Flan-T5-base} & NW   & 41.78 & 19.57 & 38.66 &  & \multirow{6}{*}{Flan-T5-base} & NW       & 49.55 \\
&  & OW (Hard) & 24.47 & 5.60 & 22.48  &  &  & OW (Hard) & 5.35 \tcbhighmath[colback=LightOrange]{\downarrow 89.2\%} \\
&  & SW (Hard) & \textbf{41.80} & \textbf{19.80} & \textbf{38.72} &  &  & SW (Hard) & \textbf{35.36} \tcbhighmath[colback=LightOrange]{\downarrow 28.6\%} \\
&  & OW (Soft) & 38.60 & 16.29 & 35.90 &  &   & OW (Soft)& 39.19 \tcbhighmath[colback=LightOrange]{\downarrow 20.9\%}\\
&  & SW (Soft) & \textbf{41.90} & \textbf{19.86} & \textbf{38.80} &  &   & SW (Soft) & \textbf{44.18} \tcbhighmath[colback=LightOrange]{\downarrow 10.8\%}\\ 
\midrule
\multirow{12}{*}{\textbf{XSUM}} & \multirow{6}{*}{BART-large}     & NW  & 45.25 & 22.15 & 37.03 & \multirow{12}{*}{\textbf{WebNLG}} & \multirow{6}{*}{BART-large} & NW  & 57.18  \\
&   & OW (Hard)& 29.60 & 7.15 & 20.83 &  &  & OW (Hard) & 9.25 \tcbhighmath[colback=LightOrange]{\downarrow 83.8\%} \\
&   & SW (Hard) & \textbf{42.44} & \textbf{18.64} & \textbf{33.91} &  &    & SW (Hard) & \textbf{48.02} \tcbhighmath[colback=LightOrange]{\downarrow 16.0\%} \\
&   & OW (Soft)& 40.07 & 16.51 & 31.50 &  &  & OW (Soft) & 44.58 \tcbhighmath[colback=LightOrange]{\downarrow 22.1\%}\\
&   & SW (Soft) & \textbf{43.83} & \textbf{20.39} & \textbf{35.42} &  &     & SW (Soft) & \textbf{52.50} \tcbhighmath[colback=LightOrange]{\downarrow 8.2\%} \\ \cmidrule{2-6} \cmidrule{8-10} 
& \multirow{6}{*}{Flan-T5-base} & NW      & 39.51 & 16.92 & 31.90 &  & \multirow{6}{*}{Flan-T5-base} & NW       & 59.77 \\
& & OW (Hard) & 22.98 & 4.80  & 16.66 &  &     & OW (Hard) & 1.80 \tcbhighmath[colback=LightOrange]{\downarrow 97.0\%} \\
& & SW (Hard) & \textbf{37.67} & \textbf{14.69} & \textbf{29.94} &  &     & SW (Hard) & \textbf{40.89} \tcbhighmath[colback=LightOrange]{\downarrow 31.6\%}\\
& & OW (Soft) & 35.23 & 12.58 & 27.52 &  &    & OW (Soft)& 45.42 \tcbhighmath[colback=LightOrange]{\downarrow 24.0\%}\\
& & SW (Soft) & \textbf{38.79}  & \textbf{15.91} & \textbf{31.03} &  &    & SW (Soft) & \textbf{53.27} \tcbhighmath[colback=LightOrange]{\downarrow 10.9\%} \\ 
\bottomrule
\end{tabular}}
\caption{Main results of comparing different watermarking strategies across various datasets and models. NW (no watermark) serves as the baseline, and adding a watermark is expected to decrease performance to trade-off detection. OW (original watermark) denotes the use of the Soft or Hard watermark \citep{kirchenbauer2023watermark} with hyperparameters $\gamma=0.5$ and $\delta \in \{2,10\}$. Our proposed SW (semantic-aware watermark) approach employs semantically related tokens to partition the green and red lists, with hyperparameters $k=1/2/5/10$, while keeping the same values of $\gamma$ and $\delta$ to ensure a fair comparison.}

\label{main_result}
\end{table*}

We extend the constraints to prioritize the inclusion of content closely related to the input within the partitioned green list, as detailed in Algorithm 1. This strategy effectively minimizes the impact of random vocabulary partitioning on the quality of generated results. The decision to utilize model embeddings to acquire semantically related tokens -- steps 1\&2 in Algorithm \ref{main_algorithm} --  is motivated by the following reasons:

\begin{itemize}
    \item Semantic Relevance: By exploiting model embeddings, we capture semantic token relationships. This ensures coherent and semantically consistent text generation by identifying tokens closely linked to the input.
    \item Enhanced Output Quality: Including semantically related tokens in the green list elevates the relevance and quality of the generated text, aligning it more effectively with the CTG task objectives.
\end{itemize}

Assume the word embeddings for a specific model have a size of $|V| \times d_{\text{emb}}$, where $|V|$ and $d_{\text{emb}}$ denote the vocabulary size and the dimension of the model's embeddings, respectively. Each row in this embedding matrix contains the representation of a particular indexed token. For each pair of token representations, we can calculate the embedding similarity using measures such as cosine similarity. This process allows us to construct a similarity matrix $\mathbf{M}$ of size $|V| \times |V|$ with token indices sorted based on their similarity values with respect to the token indexed at each row.



In our proposed semantic-aware watermarking approach, before partitioning the green list, we utilize the input context tokens as pivot points for the green list and leverage the similarity matrix $\mathbf{M}$. By combining this similarity matrix with a hyperparameter $k$, we identify the top $k$ semantically related tokens for each input token. These semantically related tokens are then included in the green list, while the remaining portion of the vocabulary is randomly partitioned.  This partitioning is carried out based on the mathematical equation presented in step 8 of Algorithm \ref{main_algorithm}.

\section{Experiments and Results}

This section provides  an overview of the datasets and models utilized in the experiments. We also present the main experimental results, including both automatic and human evaluations.

\subsection{Datasets and Models}

We conducted experiments to assess the generalization ability of our proposed method by utilizing models with different parameter sizes and architectures, including BART-base, BART-large \citep{lewis-etal-2020-bart}, Flan-T5-small, and Flan-T5-base \citep{chung2022scaling}. Our focus was on two distinct conditional text generation tasks: summarization - CNN/DailyMail \citep{see-etal-2017-get} and XSUM \citep{narayan-etal-2018-dont},  and data-to-text generation - DART \cite{nan-etal-2021-dart} and WebNLG \citep{gardent-etal-2017-webnlg}. These datasets are widely recognized for evaluating text summarization and data-to-text generation models, respectively. By conducting comprehensive evaluations across multiple datasets, tasks, and models, our objective was to thoroughly compare the differences between the original watermarking algorithm \citep{kirchenbauer2023watermark} and our proposed semantic-aware watermarking approach.

\subsection{Main Results}
Our main experimental results are presented in Table \ref{main_result}. The summarization task was evaluated using the ROUGE metric \citep{lin-2004-rouge}, while the data-to-text generation task was evaluated using BLEU \citep{papineni-etal-2002-bleu}. The table illustrates the performance of the models under various watermarking methods, highlighting the enhancements achieved by incorporating semantic constraints in watermarking for both the summarization and data-to-text generation tasks. Our proposed semantic-aware watermarking method exhibits significant improvements in comparison to the original watermarking method across all datasets and models. 

Additionally, we observe that hard watermarks invariably cause a greater decline in CTG performance compared to soft watermarks (especially ROUGE-2 for summarization and BLEU for data-to-text generation). The hard watermarks designed for language models \cite{kirchenbauer2023watermark} essentially completely forbid generation from the red list that might contain key input context, potentially leading to near-ineffective generations with almost no overlap with the reference generations.  For example, in the data-to-text generation task, the original hard watermarking method adversely affects Flan-T5-small's performance on WebNLG, resulting in a decrease of over 57.97 BLEU points with 97.0\% of performance drop. In contrast, our semantic-aware watermark effectively mitigates the impact of adding the watermark, demonstrating an 39.09 BLEU point increase over the original watermark with a performance improvement of 21.67 times.

More notably, on the CNN/DailyMail dataset, our semantic-aware watermarking method applied to the Flan-T5-base models not only mitigates the drawbacks of watermark injection but also surpasses the performance of the original generation without watermark. This can be credited to the nature of the summarization task, where a considerable amount of the target information is already present in the input. The semantic-aware watermarking method enhances the generation process by effectively harnessing this input, enabling it to capture the essential details for creating high-quality summaries. This synergy between input and target data contributes to the superior performance of the Flan-T5-small and Flan-T5-base models when utilizing the semantic-aware watermarking method in summarization tasks.

    \begin{table}[t]
    \centering
    \resizebox{\columnwidth}{!}{%
    \begin{tabular}{l c c c|c}
    \toprule
     SW (ours) vs. OW & Judge 1 &   Judge 2 & Judge 3 &Avg.\\ 
    \midrule
    SW (ours) preferred & 58\% & 54\% & 54\% & 55.33\% \\ 
    \bottomrule
    \end{tabular}
    }
\caption{ Human evaluation results on 100 randomly sampled examples, accompanied by generations from BART-base with original soft or semantic-aware watermarks, presented in a random and anonymized order.   Each example was independently annotated by three annotators, resulting in an average pairwise inter-annotator agreement of 63.33\%.}
\label{tab:human_eval}
\end{table}

\paragraph{Human Evaluation}
\label{sub-sec:human_eval_result}

In addition, we conducted a human evaluation comparing BART-base with the original and our proposed watermarks on the XSUM dataset. The human judges\footnote{All judges are native English speakers with a minimum of a bachelor's degree and were compensated at a rate of \$19.5/h.} were presented with reference summaries and generations from different watermarking algorithms in a random and anonymized order. The judges were asked to evaluate which system's summary was better and more similar to the reference. They were instructed to read the source article only when they were unable to decide or needed additional information\footnote{We made the decision to make reading the source article optional for the judges in order to prevent creating a significant cognitive burden and to encourage them to take shortcuts.}. 

\begin{figure}[t]
    \centering
    \includegraphics[scale=0.54]{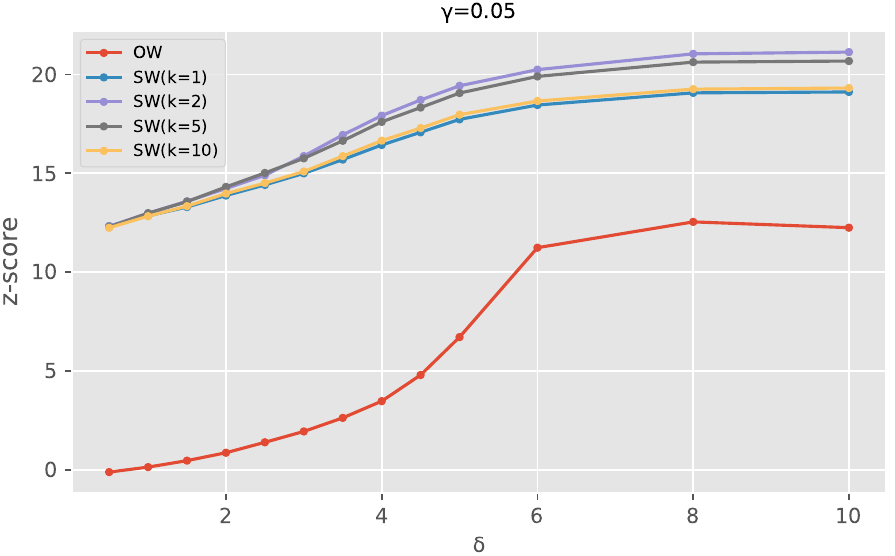}
    \caption{Watermark detection: average $z$-score under different $\delta$ settings (x-axis). Higher $z$-scores indicate stronger watermark detection confidence.  We can see that hard watermarks (greater $\delta$) are easier to detect but lead to a more significant decline in CTG performance. }
    \label{zscore-figure}
\end{figure}

\begin{figure}
    \centering
    \includegraphics[scale=0.54]{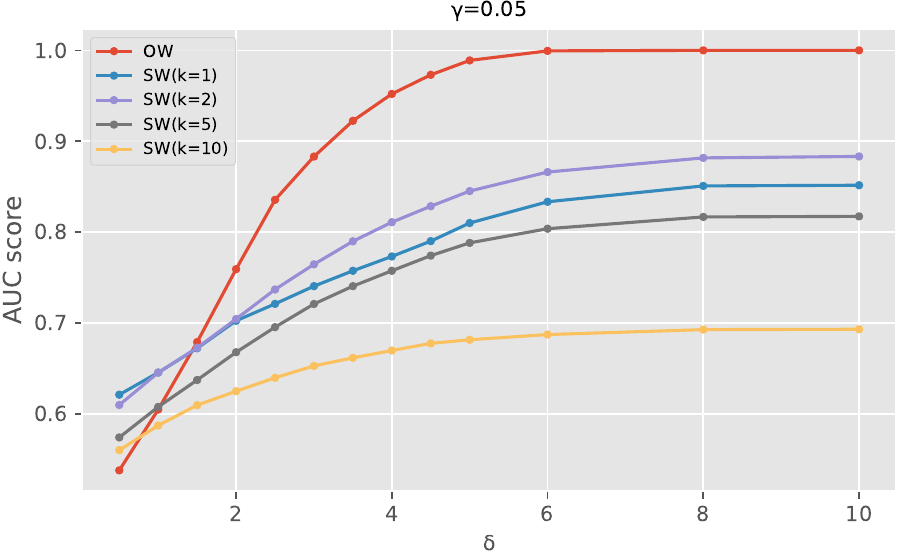}
    \caption{Watermark detection: AUC scores under different $\delta$ settings. Higher AUC scores indicates a better detection performances.}
    \label{aucscore-figure}
\end{figure}

 \begin{figure*}[!t]
 \centering
\includegraphics[scale=0.7]{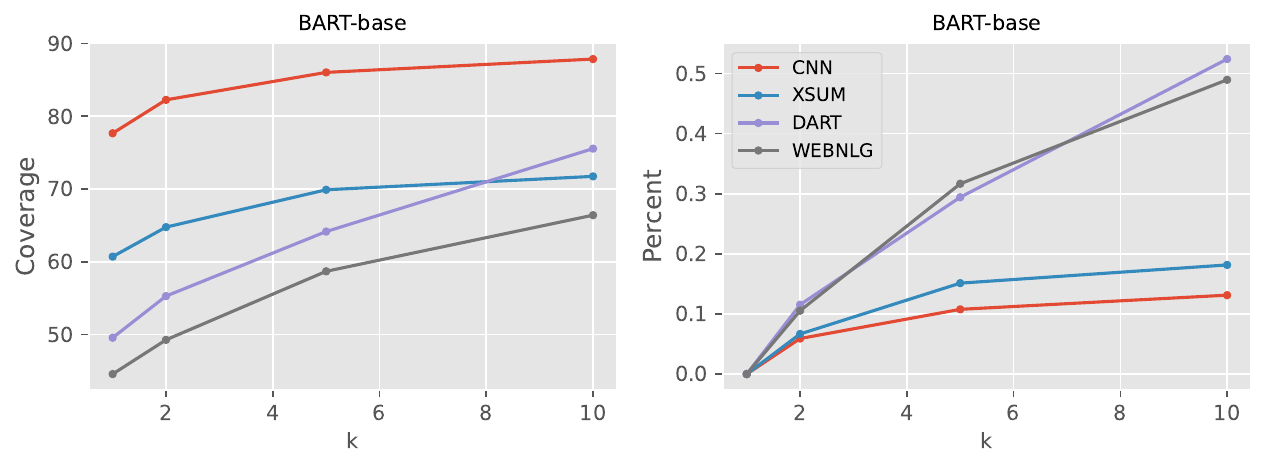}

    \caption{The coverage of target tokens by semantically related tokens varies with different datasets and values of the hyperparameter $k$ on BART-base. Increasing the value of $k$ improves the coverage of semantic tokens, aligning with our objective and motivation.}
    \label{coverage-figure}
\end{figure*}

Table~\ref{tab:human_eval} presents the results of the human evaluation. With a confidence level of 95\% and one-sided A/B tests, the semantic-aware watermark exhibits a significantly higher preference according to human judges ($p=0.0358$). Specifically, the preference for the semantic-aware watermark (55.33\%) surpasses that of the original watermark (44.67\%) by a substantial margin of 10.66\%. Moreover, pairwise inter-annotator agreement was assessed, resulting in agreement percentages of 70\%, 66\%, and 54\% for the respective evaluations. These findings strongly support the effectiveness of the semantic-aware watermarking method, highlighting its ability to enhance the quality of summarization outputs.

\subsection{Watermark Strength and Detection}

To evaluate the quality of watermarking for detection, we followed established research \cite{kirchenbauer2023watermark, yang2023watermarking} and assessed the strength using the average $z$-score and the area under the curve (AUC) score. Figure \ref{zscore-figure} and Figure \ref{aucscore-figure} present the $z$-score and AUC results, respecively. 

A higher $z$-score generally indicates a greater presence of tokens from the ``green list" in the generated results, increasing the likelihood of successful detection. However, in the context of conditional text generation tasks, maintaining consistency in the length of the generated results with the original model is crucial. It has been observed that the $z$-score tends to increase with the length of the generated text \cite{kirchenbauer2023watermark}. To address this, we introduce an additional penalty term to the $z$-score, incorporating the ratio of the average length of the generated results to the average length of the original model's output without the watermark.

As seen in Figure \ref{zscore-figure}, the semantic-aware watermarking method significantly outperforms its counterpart in terms of $z$-score, reflecting a higher inclusion of ``green list" tokens in the generated output. Under normal circumstances (e.g., language modeling), a higher average $z$-score indicates stronger detectability \cite{kirchenbauer2023watermark}. However, as Figure \ref{aucscore-figure} illustrates, the AUC curve for the original watermarking method surpasses ours, as our constructed green lists incorporate more input tokens that humans would commonly use. Consequently, human-generated text also contains more green list tokens.

The disparity between the $z$-scores and AUC scores of semantic-aware watermarks highlights an additional challenge in applying watermarks for CTG: \textbf{the common human practice of utilizing input-similar tokens in CTG introduces complexity to the watermark detection process}. Our method, despite showing remarkable improvements in ROUGE or BLEU metrics and hence bearing closer resemblance to the reference, contributes to a slight dip in the final AUC scores. This scenario indicates a trade-off between enhancing the ROUGE or BLEU scores, indicative of increased similarity to the reference, and preserving detectability.  Notwithstanding this, our empirical results compellingly argue that the significant rise in performance (up to $\sim 2167\%$) outweighs the detection decreases (Avg. $\sim 12.6\%$); further increasing this advantage margin remains an area for future exploration.

\section{Analysis}
This section analyzes the hyperparameters, focusing on: $k$, introduced by our semantic watermark; $\gamma$ and $\delta$, inherited from \citet{kirchenbauer2023watermark}.

\subsection{Semantic $k$ Analysis}
\begin{table}[ht]
\centering

\resizebox{0.4\textwidth}{!}{\begin{tabular}{lcccc}
\toprule
Method & \multicolumn{3}{c}{BLEU} \\ \cmidrule{2-4}
$\gamma$ & $\gamma=0.25$ & $\gamma=0.5$ & $\gamma=0.75$\\
\midrule

NW & 45.90 & - & - \\ 
\midrule
OW & 37.32 & 35.99 & 39.01 \\ 
SW (k=1) & 37.23 & 38.46 & 41.36 \\
SW (k=2) & 38.10 & 39.29 & 42.01 \\
SW (k=5) & 38.87 & 38.63 & 42.24 \\
SW (k=10) & \textbf{41.37} & \textbf{42.89} & \textbf{44.59} \\
\bottomrule
\end{tabular}}
\caption{The effect of the hyperparameter $k$ on the results of the DART dataset using the BART-base with $\gamma \in \{0.25, 0.5, 0.75\}$ and $\delta=2$.}
\label{k-table} 
\end{table}

The semantic-aware watermark uses a hyperparameter, $k$, to determine the extent of semantically related tokens, derived from word embedding similarities during decoding, that are integrated into the green list. Table \ref{k-table} shows that \textbf{increasing  $k$ in semantic-aware watermarks improve the CTG performance}.  We hypothesize that this improvement stems from that increasing $k$ includes more reference tokens in the green list, leading to a broader coverage of tokens that humans typically use for CTG generation.

To validate our hypothesis and study the relationship between $k$ and target token coverage, we carried out experiments by measuring the overlaps between semantically related tokens and the reference target tokens under different $k$ values. Figure \ref{coverage-figure} (left) presents curves, which, with increasing $k$, demonstrate a correlation with an increased proportion of target unigram text tokens covered by semantically related tokens.

Interestingly, when we adjust the setup to measure the relative percentage of coverage increase with higher $k$ values, we observe different trends for various CTG tasks. Figure \ref{coverage-figure} (right) indicates that watermarks with larger $k$ values have a more significant performance improvement impact on data-to-text generation tasks compared to summarization tasks. This observation is also reflected in the findings that an increased $k$ leads to substantial improvements in BLEU scores for data-to-text generation, compared to the ROUGE score improvements for summarization (More details in Appendix). Specifically, DART and WEBNLG show greater sensitivity to $k$, where its increase yields better results.

\subsection{$\gamma$ and $\delta$ Analysis}
\begin{figure}[t]
    \centering
    \includegraphics[scale=0.64]{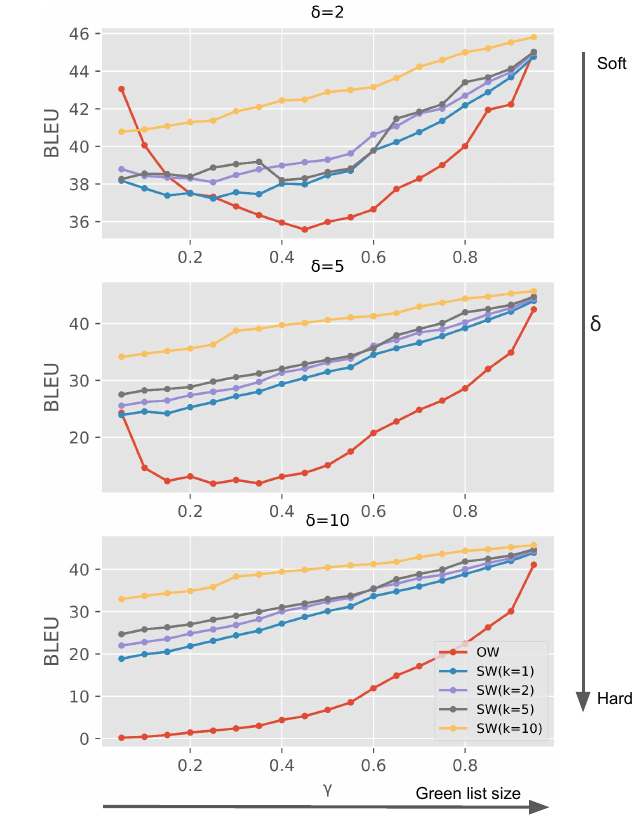}
    \caption{The impact of $\gamma$ on DART results with  settings of $\delta=2/5/10$. $\gamma$ controls the size of the green list. From $\delta=2$ to $\delta=5$, the watermarking method tends to change from a soft watermark to a hard watermark, and the probability of generating tokens from the green list gradually increases.}
    \label{gamma-figure}
\end{figure}

The soft watermarking method \citep{kirchenbauer2023watermark} depends on two hyperparameters: $\gamma$ and $\delta$. $\gamma$ regulates the size of the green list during partitioning, whereas $\delta$ dictates the intensity of watermarks applied to the logits of green list tokens. Essentially, a very large $\delta$ (e.g., 10) is equivalent to the hard watermarks that entirely prohibits tokens from the red list from being generated. This section compares original and semantic-aware watermarks under varying $\gamma$ and $\delta$ values, demonstrating that our proposed watermark consistently outperforms the original across different hyperparameter settings.

Increasing $\gamma$ incorporates more words into the green list, typically lessening the watermark's impact on model performance. Surprisingly, Table \ref{k-table} shows that the original watermarking method performs poorly when $\gamma = 0.5$. To further explore possible reasons for this and to test our method under different setups, we conducted a comparative analysis with varying $\gamma$ and $\delta$ set to 2, 5, and 10. Figure \ref{gamma-figure} indicates that the semantic-aware watermark \textbf{consistently} outperforms the original watermark, except when $\delta$ is set to 2 with relatively small $\gamma$ values. Decreasing $\gamma$ reduces the number of selected and enhanced tokens due to the smaller green list size. As a result, the model's performance is expected to gradually decrease with a smaller watermark. However, the change curve of the original method in the $\gamma < 0.2$ (when $\delta$=2) range deviates from these expectations.

\begin{figure}[t]
    \centering
    \includegraphics[scale=0.5]{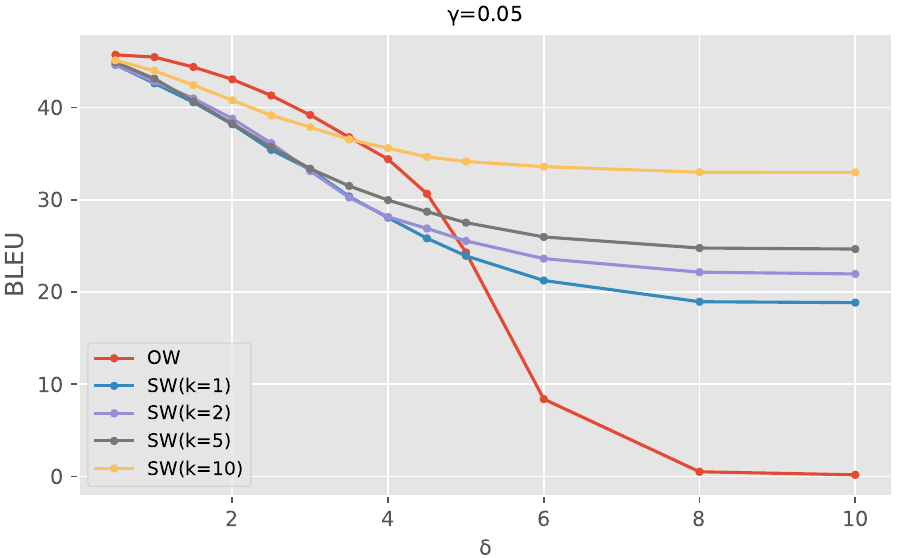}
    \caption{The impact of $\delta$, which controls the extent of enhancement applied to the logits, on the DART results.  }
    \label{delta-figure}
\end{figure}

We hypothesize that this irregularity arises from the negligible impact of soft watermark when $\gamma$ is small. This happens when soft watermarks with an extremely small green list scarcely affect logits predictions. To confirm this, we examined the impact of varying $\delta$ on the BART-base model's performance using the DART dataset under extrem small $\gamma$, as shown in Figure \ref{delta-figure}. We observe that when $\gamma$ is set extremely low ($\gamma=0.05$) in the soft watermark settings (i.e., $\delta \leq 4$), there is hardly any performance trade-off upon adding watermarks, suggesting ineffective watermarks for detection.

In addition, to ensure that semantically related tokens included in the green list for the semantic-aware watermark do not exceed the green list size, especially the ones obtained with a large $k$, we calculate the percentage of these semantically related tokens relative to the overall vocabulary size. Table \ref{target-percent-table} reveals that it is significantly lower than the green list size dictated by $\gamma$.

\begin{table}[t]
\resizebox{0.45\textwidth}{!}{
\begin{tabular}{c|cccc}
\diagbox[width=6em]{Dataset}{\\$k$}     & 1      & 2      & 5      & 10     \\ \toprule
DART   & 0.0004 & 0.0009 & 0.0020 & 0.0037 \\ 
WebNLG & 0.0005 & 0.0009 & 0.0022 & 0.0039 \\ 
\end{tabular}}
\caption{The percentage of semantically related tokens to the size of the vocabulary $V$.}
\label{target-percent-table}
\end{table}

\section{Conclusion}

Our study reveals a significant performance drop when random watermarks are directly applied to conditional text generation tasks without considering the task-specific context. To tackle this challenge, we propose a semantic-aware watermarking algorithm that incorporates hash function and carefully takes into account the input context of conditional generation tasks. We extensively evaluated our method on diverse datasets and models, including summarization, data-to-text generation, and various text generation models like BART and Flan-T5. The results demonstrate that our proposed method effectively mitigates the quality degradation associated with watermarking techniques, as confirmed by both automatic and human evaluations. These findings emphasize the importance of task-specific approaches when applying watermarking methods to ensure optimal performance in conditional text generation tasks.

\bibliography{aaai24, anthology}
\clearpage

\section{Appendix}
\label{sec:appendix}

\subsection{Comprehensive Results}
In this section, we furnish a detailed analysis of our experimental outcomes to augment the findings presented in Table \ref{main_result}, Figure \ref{aucscore-figure}, Figure \ref{zscore-figure}, Figure \ref{coverage-figure},  Figure \ref{delta-figure}. These encompass comparative evaluations conducted across diverse models and datasets, under a variety of experimental conditions.

\begin{figure}[h]
 \centering 
\includegraphics[scale=0.36]{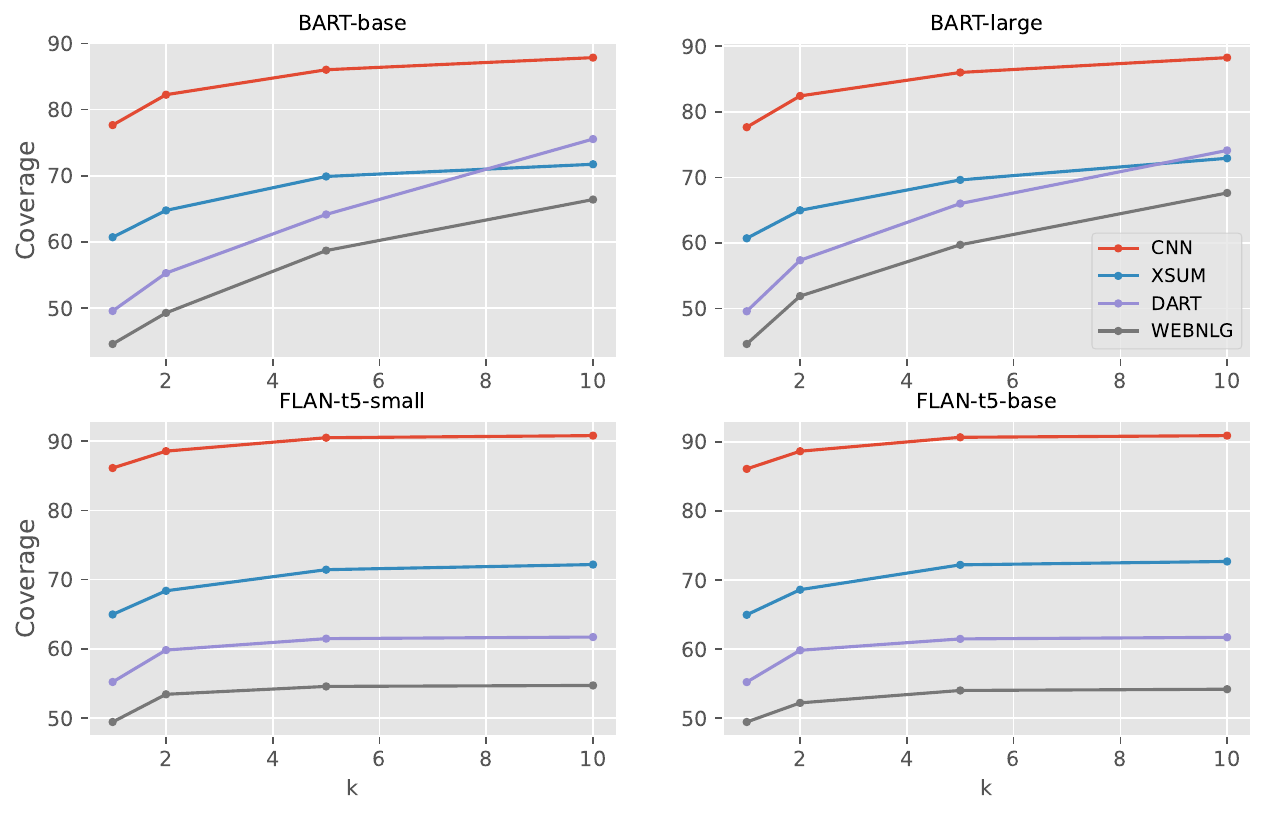}
\includegraphics[scale=0.36]{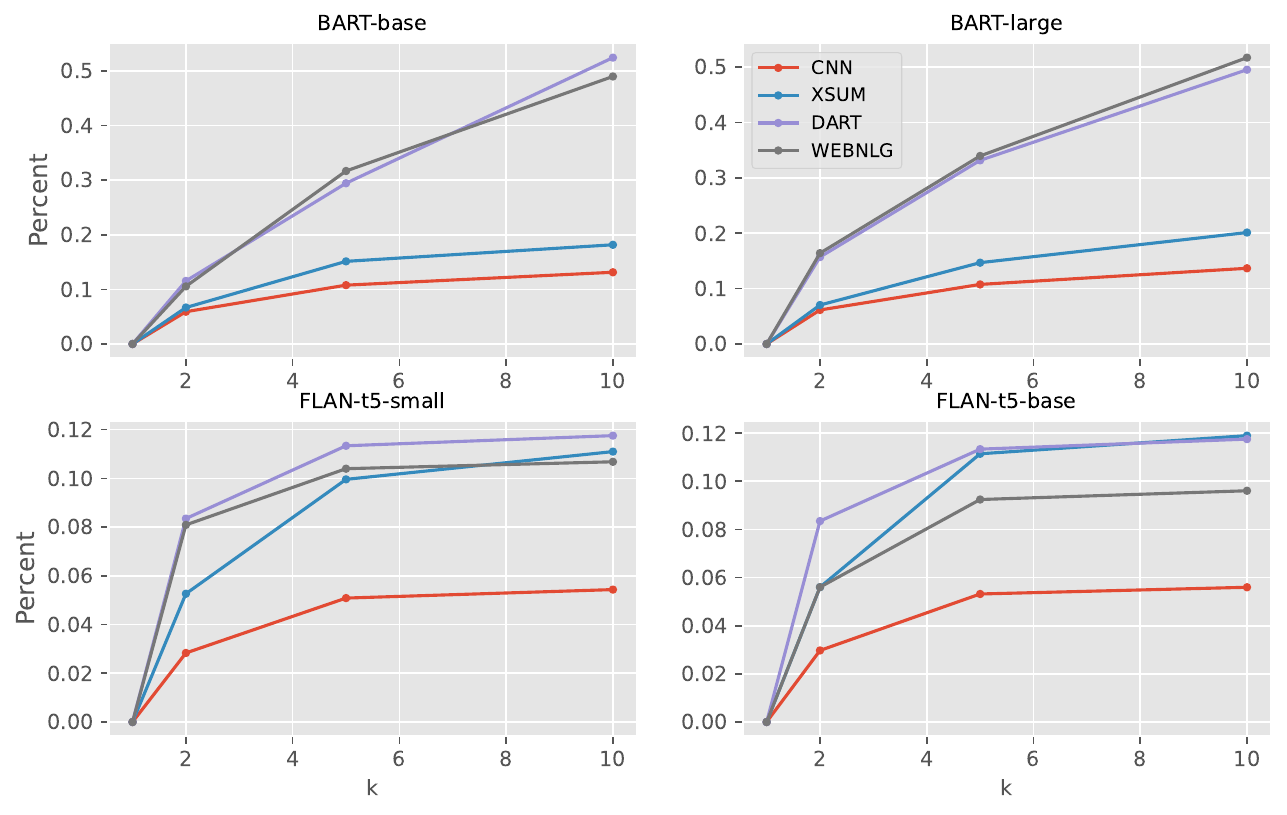}
    \caption{The coverage of target tokens by semantically related tokens varies with different datasets and values of the hyperparameter $k$. Increasing the value of $k$ improves the coverage of semantic tokens, aligning with our objective and motivation.}
    \label{coverage-figure-full}
\end{figure}

\begin{figure}[t]
    \centering
    \includegraphics[scale=0.37]{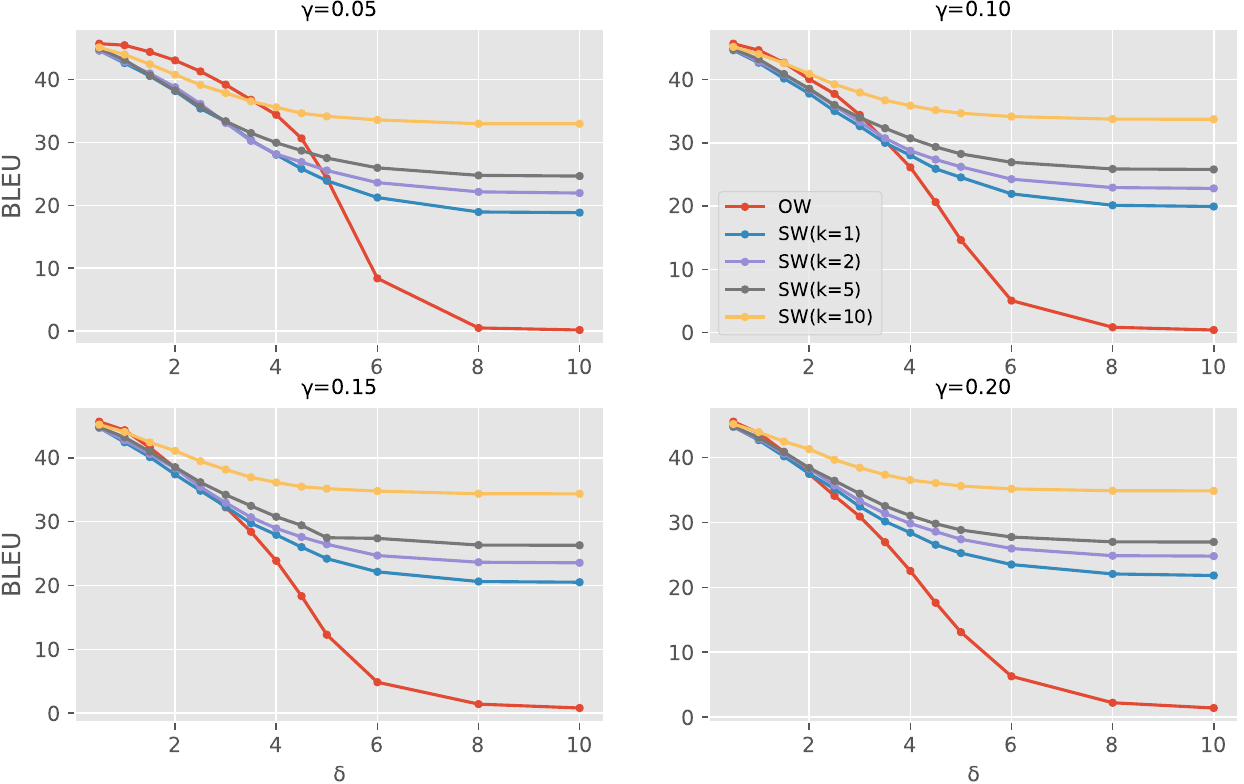}
    \caption{The impact of $\delta$, which controls the extent of enhancement applied to the logits, on the DART results.  }
    \label{delta-figure-full}

\end{figure}

\begin{figure}[t]
    \centering
    \includegraphics[scale=0.37]{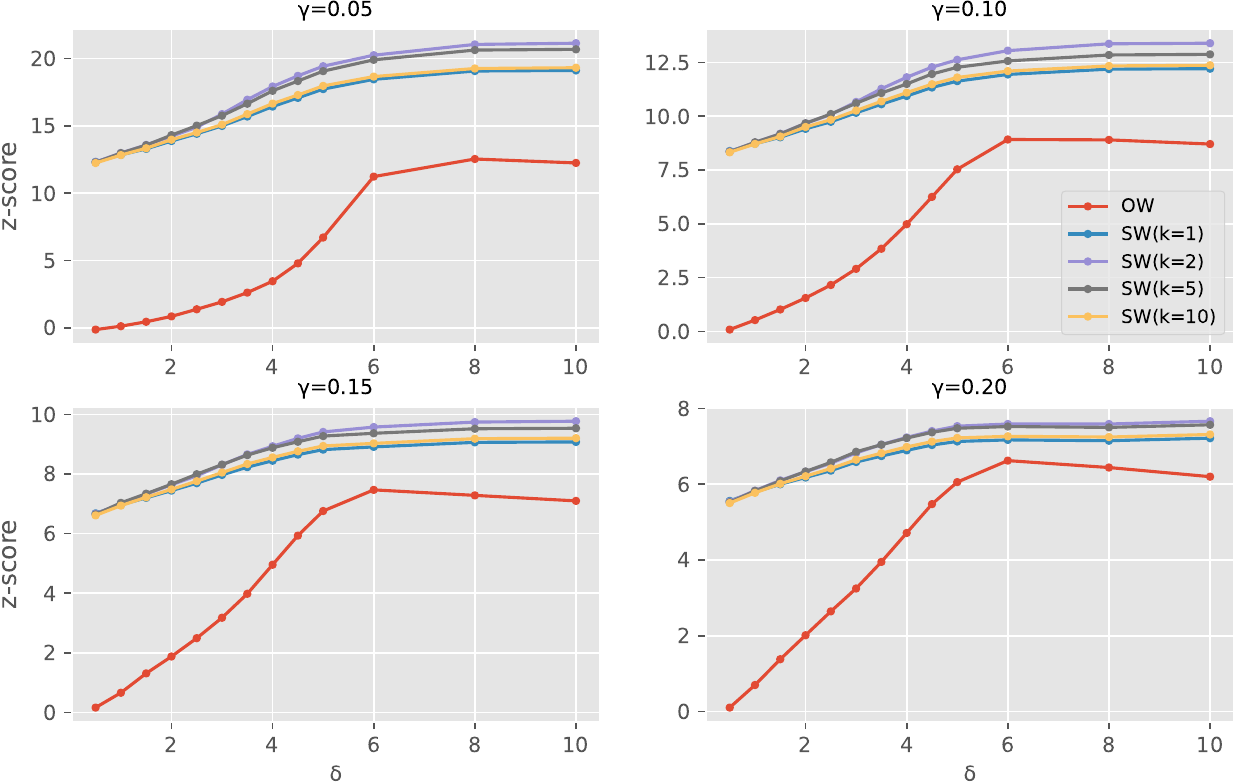}
    \caption{Watermark detection: average $z$-score under different $\delta$ setting.}
    \label{zscore-figure-full}
\end{figure}

\begin{figure}
    \centering
    \includegraphics[scale=0.37]{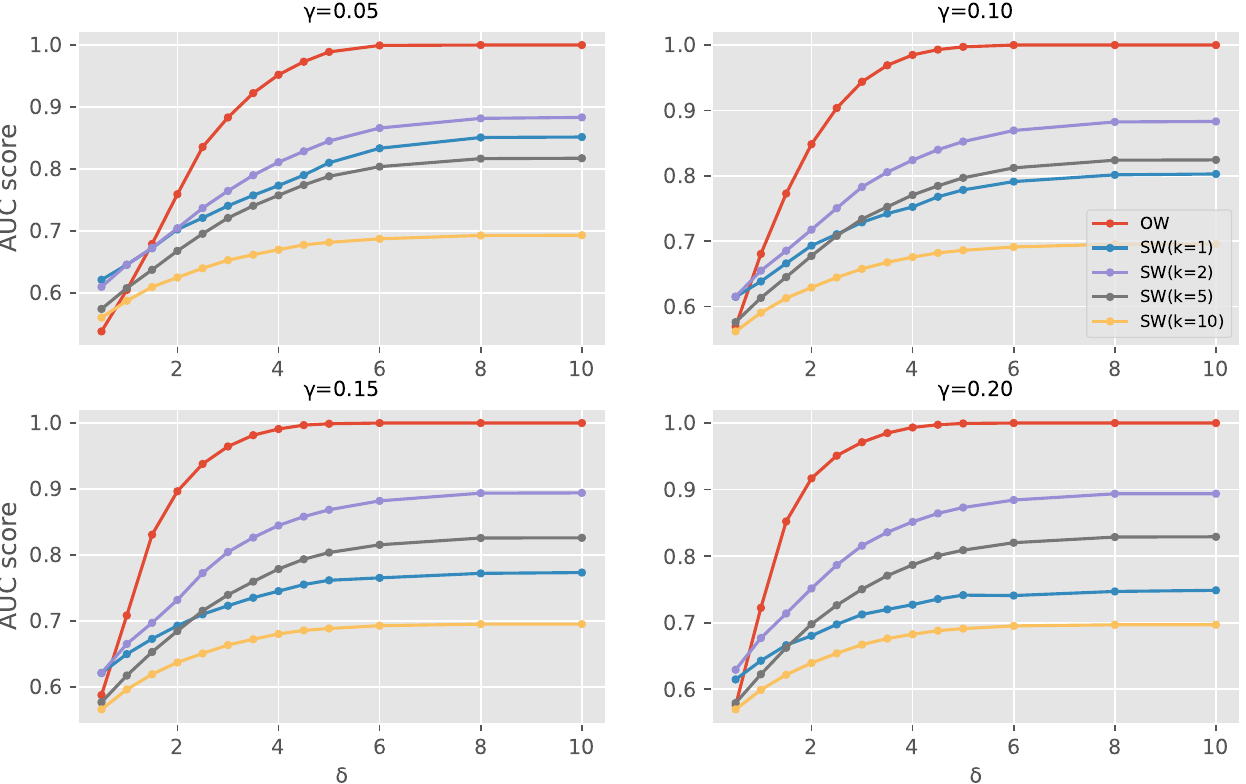}
    \caption{Watermark detection: AUC scores under different $\delta$ settings are presented. A higher AUC score indicates a better detection performance.}
    \label{aucscore-figure-full}
\end{figure}

\begin{table*}[t]
\resizebox{\textwidth}{!}{

\begin{tabular}{lllllll|lllll}
\hline
Dataset               & model                           & method   & $\gamma$ and $k$     & ROUGE-1 & ROUGE-2 & ROUGE-L & Dataset                & model                           & method   & $\gamma$ and $k$     & BLEU  \\ \hline
\multirow{64}{*}{CNN} & \multirow{16}{*}{BART-base}     & NW       & -                    & 42.02   & 19.46   & 39.04   & \multirow{64}{*}{DART} & \multirow{16}{*}{BART-base}     & NW       & -                    & 45.90 \\ \cline{3-7} \cline{10-12} 
                      &                                 & OW       & $\gamma=0.25$       & 39.65   & 17.28   & 36.84   &                        &                                 & OW       & $\gamma=0.25$       & 37.32 \\
                      &                                 & SW (ours) & $\gamma=0.25 + k=1$  & 40.93   & 18.81   & 37.68   &                        &                                 & SW (ours) & $\gamma=0.25+k=1$  & 37.23 \\
                      &                                 & SW (ours) & $\gamma=0.25+k=2$  & 41.34   & 19.12   & 38.20   &                        &                                 & SW (ours) & $\gamma=0.25+k=2$  & 38.10 \\
                      &                                 & SW (ours) & $\gamma=0.25+k=5$  & 41.36   & 19.14   & 38.21   &                        &                                 & SW (ours) & $\gamma=0.25+k=5$  & 38.87 \\
                      &                                 & SW (ours) & $\gamma=0.25+k=10$ & 41.45   & 19.22   & 38.28   &                        &                                 & SW (ours) & $\gamma=0.25+k=10$ & 41.37 \\ \cline{3-7} \cline{10-12} 
                      &                                 & OW       & $\gamma=0.5$        & 38.13   & 16.56   & 35.33   &                        &                                 & OW       & $\gamma=0.5$        & 35.99 \\
                      &                                 & SW (ours) & $\gamma=0.5+k=1$   & 40.93   & 18.81   & 37.68   &                        &                                 & SW (ours) & $\gamma=0.5+k=1$   & 38.46 \\
                      &                                 & SW (ours) & $\gamma=0.5+k=2$   & 41.46   & 19.17   & 38.37   &                        &                                 & SW (ours) & $\gamma=0.5+k=2$   & 39.29 \\
                      &                                 & SW (ours) & $\gamma=0.5+k=5$   & 41.65   & 19.30   & 38.54   &                        &                                 & SW (ours) & $\gamma=0.5+k=5$   & 38.36 \\
                      &                                 & SW (ours) & $\gamma=0.5+k=10$  & 41.59   & 19.28   & 38.49   &                        &                                 & SW (ours) & $\gamma=0.5+k=10$  & 42.89 \\ \cline{3-7} \cline{10-12} 
                      &                                 & OW       & $\gamma=0.75$       & 40.47   & 18.08   & 37.71   &                        &                                 & OW       & $\gamma=0.75$       & 39.01 \\
                      &                                 & SW (ours) & $\gamma=0.75+k=1$  & 41.82   & 19.33   & 38.73   &                        &                                 & SW (ours) & $\gamma=0.75+k=1$  & 41.36 \\
                      &                                 & SW (ours) & $\gamma=0.75+k=2$  & 41.73   & 19.29   & 29.01   &                        &                                 & SW (ours) & $\gamma=0.75+k=2$  & 42.01 \\
                      &                                 & SW (ours) & $\gamma=0.75+k=5$  & 41.78   & 19.34   & 38.75   &                        &                                 & SW (ours) & $\gamma=0.75+k=5$  & 42.24 \\
                      &                                 & SW (ours) & $\gamma=0.75+k=10$ & 41.84   & 19.38   & 38.80   &                        &                                 & SW (ours) & $\gamma=0.75+k=10$ & 44.59 \\ \cline{2-7} \cline{9-12} 
                      & \multirow{16}{*}{BART-large}    & NW       & -                    & 43.80   & 20.88   & 40.73   &                        & \multirow{16}{*}{BART-large}    & NW       & -                    & 47.78 \\ \cline{3-7} \cline{10-12} 
                      &                                 & OW       & $\gamma=0.25$       & 42.17   & 18.40   & 39.26   &                        &                                 & OW       & $\gamma=0.25$       & 38.66 \\
                      &                                 & SW (ours) & $\gamma=0.25+k=1$  & 43.31   & 20.61   & 40.25   &                        &                                 & SW (ours) & $\gamma=0.25+k=1$  & 38.08 \\
                      &                                 & SW (ours) & $\gamma=0.25+k=2$  & 43.35   & 20.68   & 40.31   &                        &                                 & SW (ours) & $\gamma=0.25+k=2$  & 39.24 \\
                      &                                 & SW (ours) & $\gamma=0.25+k=5$  & 43.41   & 20.73   & 40.38   &                        &                                 & SW (ours) & $\gamma=0.25+k=5$  & 39.71 \\
                      &                                 & SW (ours) & $\gamma=0.25+k=10$ & 43.50   & 20.82   & 40.50   &                        &                                 & SW (ours) & $\gamma=0.25+k=10$ & 41.81 \\ \cline{3-7} \cline{10-12} 
                      &                                 & OW       & $\gamma=0.5$        & 42.46   & 18.33   & 39.52   &                        &                                 & OW       & $\gamma=0.5$        & 37.07 \\
                      &                                 & SW (ours) & $\gamma=0.5+k=1$   & 43.38   & 20.71   & 40.34   &                        &                                 & SW (ours) & $\gamma=0.5+k=1$   & 39.63 \\
                      &                                 & SW (ours) & $\gamma=0.5+k=2$   & 43.50   & 20.83   & 40.62   &                        &                                 & SW (ours) & $\gamma=0.5+k=2$   & 40.26 \\
                      &                                 & SW (ours) & $\gamma=0.5+k=5$   & 43.49   & 20.81   & 40.47   &                        &                                 & SW (ours) & $\gamma=0.5+k=5$   & 42.03 \\
                      &                                 & SW (ours) & $\gamma=0.5+k=10$  & 43.50   & 20.82   & 40.50   &                        &                                 & SW (ours) & $\gamma=0.5+k=10$  & 44.04 \\ \cline{3-7} \cline{10-12} 
                      &                                 & OW       & $\gamma=0.75$       & 43.13   & 19.55   & 40.16   &                        &                                 & OW       & $\gamma=0.75$       & 40.69 \\
                      &                                 & SW (ours) & $\gamma=0.75+k=1$  & 43.46   & 20.76   & 40.43   &                        &                                 & SW (ours) & $\gamma=0.75+k=1$  & 42.71 \\
                      &                                 & SW (ours) & $\gamma=0.75+k=2$  & 43.46   & 20.76   & 40.51   &                        &                                 & SW (ours) & $\gamma=0.75+k=2$  & 44.27 \\
                      &                                 & SW (ours) & $\gamma=0.75+k=5$  & 43.57   & 20.88   & 40.56   &                        &                                 & SW (ours) & $\gamma=0.75+k=5$  & 44.22 \\
                      &                                 & SW (ours) & $\gamma=0.75+k=10$ & 43.53   & 20.87   & 40.54   &                        &                                 & SW (ours) & $\gamma=0.75+k=10$ & 45.75 \\ \cline{2-7} \cline{9-12} 
                      & \multirow{16}{*}{Flan-T5-small} & NW       & -                    & 38.96   & 17.35   & 35.84   &                        & \multirow{16}{*}{Flan-T5-small} & NW       & -                    & 47.99 \\ \cline{3-7} \cline{10-12} 
                      &                                 & OW       & $\gamma=0.25$       & 35.44   & 14.46   & 32.84   &                        &                                 & OW       & $\gamma=0.25$       & 37.47 \\
                      &                                 & SW (ours) & $\gamma=0.25+k=1$  & 39.75   & 17.86   & 36.56   &                        &                                 & SW (ours) & $\gamma=0.25+k=1$  & 40.70 \\
                      &                                 & SW (ours) & $\gamma=0.25+k=2$  & 39.86   & 17.96   & 36.68   &                        &                                 & SW (ours) & $\gamma=0.25+k=2$  & 40.77 \\
                      &                                 & SW (ours) & $\gamma=0.25+k=5$  & 39.82   & 17.91   & 36.64   &                        &                                 & SW (ours) & $\gamma=0.25+k=5$  & 40.63 \\
                      &                                 & SW (ours) & $\gamma=0.25+k=10$ & 39.82   & 17.91   & 36.64   &                        &                                 & SW (ours) & $\gamma=0.25+k=10$ & 40.58 \\ \cline{3-7} \cline{10-12} 
                      &                                 & OW       & $\gamma=0.5$        & 35.51   & 14.56   & 32.90   &                        &                                 & OW       & $\gamma=0.5$        & 36.32 \\
                      &                                 & SW (ours) & $\gamma=0.5+k=1$   & 39.76   & 17.88   & 36.58   &                        &                                 & SW (ours) & $\gamma=0.5+k=1$   & 41.66 \\
                      &                                 & SW (ours) & $\gamma=0.5+k=2$   & 39.86   & 17.95   & 36.68   &                        &                                 & SW (ours) & $\gamma=0.5+k=2$   & 42.52 \\
                      &                                 & SW (ours) & $\gamma=0.5+k=5$   & 39.83   & 17.91   & 36.65   &                        &                                 & SW (ours) & $\gamma=0.5+k=5$   & 42.65 \\
                      &                                 & SW (ours) & $\gamma=0.5+k=10$  & 39.80   & 17.90   & 36.63   &                        &                                 & SW (ours) & $\gamma=0.5+k=10$  & 42.61 \\ \cline{3-7} \cline{10-12} 
                      &                                 & OW       & $\gamma=0.75$       & 37.15   & 15.83   & 34.30   &                        &                                 & OW       & $\gamma=0.75$       & 39.93 \\
                      &                                 & SW (ours) & $\gamma=0.75+k=1$  & 39.81   & 17.91   & 36.61   &                        &                                 & SW (ours) & $\gamma=0.75+k=1$  & 45.16 \\
                      &                                 & SW (ours) & $\gamma=0.75+k=2$  & 39.85   & 17.94   & 36.68   &                        &                                 & SW (ours) & $\gamma=0.75+k=2$  & 45.04 \\
                      &                                 & SW (ours) & $\gamma=0.75+k=5$  & 39.83   & 17.91   & 36.66   &                        &                                 & SW (ours) & $\gamma=0.75+k=5$  & 45.01 \\
                      &                                 & SW (ours) & $\gamma=0.75+k=10$ & 39.83   & 17.92   & 36.66   &                        &                                 & SW (ours) & $\gamma=0.75+k=10$ & 45.00 \\ \cline{2-7} \cline{9-12} 
                      & \multirow{16}{*}{Flan-T5-base}  & NW       & -                    & 41.78   & 19.57   & 38.66   &                        & \multirow{16}{*}{Flan-T5-base}  & NW       & -                    & 49.55 \\ \cline{3-7} \cline{10-12} 
                      &                                 & OW       & $\gamma=0.25$       & 38.24   & 16.12   & 35.59   &                        &                                 & OW       & $\gamma=0.25$       & 38.92 \\
                      &                                 & SW (ours) & $\gamma=0.25+k=1$  & 41.70   & 19.70   & 38.60   &                        &                                 & SW (ours) & $\gamma=0.25+k=1$  & 42.06 \\
                      &                                 & SW (ours) & $\gamma=0.25+k=2$  & 41.78   & 19.77   & 38.68   &                        &                                 & SW (ours) & $\gamma=0.25+k=2$  & 42.74 \\
                      &                                 & SW (ours) & $\gamma=0.25+k=5$  & 41.88   & 19.82   & 38.78   &                        &                                 & SW (ours) & $\gamma=0.25+k=5$  & 42.87 \\
                      &                                 & SW (ours) & $\gamma=0.25+k=10$ & 41.88   & 19.84   & 38.78   &                        &                                 & SW (ours) & $\gamma=0.25+k=10$ & 42.83 \\ \cline{3-7} \cline{10-12} 
                      &                                 & OW       & $\gamma=0.5$        & 38.60   & 16.29   & 35.90   &                        &                                 & OW       & $\gamma=0.5$        & 39.13 \\
                      &                                 & SW (ours) & $\gamma=0.5+k=1$   & 41.81   & 19.80   & 38.70   &                        &                                 & SW (ours) & $\gamma=0.5+k=1$   & 43.12 \\
                      &                                 & SW (ours) & $\gamma=0.5+k=2$   & 41.87   & 19.87   & 38.79   &                        &                                 & SW (ours) & $\gamma=0.5+k=2$   & 43.64 \\
                      &                                 & SW (ours) & $\gamma=0.5+k=5$   & 41.90   & 19.86   & 38.80   &                        &                                 & SW (ours) & $\gamma=0.5+k=5$   & 44.18 \\
                      &                                 & SW (ours) & $\gamma=0.5+k=10$  & 41.88   & 19.85   & 38.80   &                        &                                 & SW (ours) & $\gamma=0.5+k=10$  & 44.06 \\ \cline{3-7} \cline{10-12} 
                      &                                 & OW       & $\gamma=0.75$       & 39.21   & 17.37   & 36.42   &                        &                                 & OW       & $\gamma=0.75$       & 41.36 \\
                      &                                 & SW (ours) & $\gamma=0.75+k=1$  & 41.88   & 19.84   & 38.78   &                        &                                 & SW (ours) & $\gamma=0.75+k=1$  & 46.43 \\
                      &                                 & SW (ours) & $\gamma=0.75+k=2$  & 41.90   & 19.89   & 38.82   &                        &                                 & SW (ours) & $\gamma=0.75+k=2$  & 46.78 \\
                      &                                 & SW (ours) & $\gamma=0.75+k=5$  & 41.91   & 19.87   & 38.83   &                        &                                 & SW (ours) & $\gamma=0.75+k=5$  & 46.67 \\
                      &                                 & SW (ours) & $\gamma=0.75+k=10$ & 41.89   & 19.86   & 38.81   &                        &                                 & SW (ours) & $\gamma=0.75+k=10$ & 46.46 \\ \hline
\end{tabular}
}
\caption{Complete results on the CNN and DART dataset.}
\end{table*}

\begin{table*}[t]
\resizebox{\textwidth}{!}{\begin{tabular}{lllllll|lllll}
\hline
Dataset                & model                           & method   & $\gamma$ and $k$     & ROUGE-1 & ROUGE-2 & ROUGE-L & Dataset                  & model                           & method   & $\gamma$ and $k$     & BLEU  \\ \hline
\multirow{64}{*}{XSUM} & \multirow{16}{*}{BART-base}     & NW       & -                    & 42.36   & 19.42   & 34.40   & \multirow{64}{*}{WebNLG} & \multirow{16}{*}{BART-base}     & NW       & -                    & 54.45 \\ \cline{3-7} \cline{10-12} 
                   &                                 & OW       & $\gamma==0.25$       & 37.84   & 14.76   & 29.89   &                          &                                 & OW       & $\gamma==0.25$       & 44.46 \\
                   &                                 & SW (ours) & $\gamma=0.25$+$k=1$  & 40.09   & 16.67   & 31.65   &                          &                                 & SW (ours) & $\gamma=0.25$+$k=1$  & 47.03 \\
                   &                                 & SW (ours) & $\gamma=0.25$+$k=2$  & 40.28   & 16.95   & 31.93   &                          &                                 & SW (ours) & $\gamma=0.25$+$k=2$  & 48.17 \\ 
                   &                                 & SW (ours) & $\gamma=0.25$+$k=5$  & 40.78   & 17.27   & 32.36   &                          &                                 & SW (ours) & $\gamma=0.25$+$k=5$  & 48.53 \\
                   &                                 & SW (ours) & $\gamma=0.25$+$k=10$ & 40.81   & 17.41   & 32.38   &                          &                                 & SW (ours) & $\gamma=0.25$+$k=10$ & 49.99 \\ \cline{3-7} \cline{10-12} 
                   &                                 & OW       & $\gamma==0.5$        & 37.99   & 14.66   & 29.82   &                          &                                 & OW       & $\gamma==0.5$        & 43.14 \\
                   &                                 & SW (ours) & $\gamma=0.5$+$k=1$   & 40.95   & 17.53   & 32.56   &                          &                                 & SW (ours) & $\gamma=0.5$+$k=1$   & 47.48 \\
                   &                                 & SW (ours) & $\gamma=0.5$+$k=2$   & 41.00   & 17.65   & 32.70   &                          &                                 & SW (ours) & $\gamma=0.5$+$k=2$   & 48.49 \\
                   &                                 & SW (ours) & $\gamma=0.5$+$k=5$   & 41.27   & 17.82   & 32.92   &                          &                                 & SW (ours) & $\gamma=0.5$+$k=5$   & 49.98 \\
                   &                                 & SW (ours) & $\gamma=0.5$+$k=10$  & 41.36   & 17.97   & 33.03   &                          &                                 & SW (ours) & $\gamma=0.5$+$k=10$  & 51.25 \\ \cline{3-7} \cline{10-12} 
                   &                                 & OW       & $\gamma==0.75$       & 40.07   & 16.67   & 31.80   &                          &                                 & OW       & $\gamma==0.75$       & 45.88 \\
                   &                                 & SW (ours) & $\gamma=0.75$+$k=1$  & 41.60   & 18.26   & 33.29   &                          &                                 & SW (ours) & $\gamma=0.75$+$k=1$  & 50.95 \\
                   &                                 & SW (ours) & $\gamma=0.75$+$k=2$  & 41.65   & 18.43   & 33.42   &                          &                                 & SW (ours) & $\gamma=0.75$+$k=2$  & 50.50 \\
                   &                                 & SW (ours) & $\gamma=0.75$+$k=5$  & 41.82   & 18.59   & 33.61   &                          &                                 & SW (ours) & $\gamma=0.75$+$k=5$  & 51.60 \\
                   &                                 & SW (ours) & $\gamma=0.75$+$k=10$ & 41.85   & 18.61   & 33.60   &                          &                                 & SW (ours) & $\gamma=0.75$+$k=10$ & 53.01 \\ \cline{2-7} \cline{9-12} 
                   & \multirow{16}{*}{BART-large}    & NW       & -                    & 45.25   & 22.15   & 37.03   &                          & \multirow{16}{*}{BART-large}    & NW       & -                    & 57.18 \\ \cline{3-7} \cline{10-12} 
                   &                                 & OW       & $\gamma==0.25$       & 40.13   & 16.75   & 31.83   &                          &                                 & OW       & $\gamma==0.25$       & 47.87 \\
                   &                                 & SW (ours) & $\gamma=0.25$+$k=1$  & 42.26   & 18.63   & 33.54   &                          &                                 & SW (ours) & $\gamma=0.25$+$k=1$  & 49.03 \\
                   &                                 & SW (ours) & $\gamma=0.25$+$k=2$  & 42.51   & 18.92   & 33.90   &                          &                                 & SW (ours) & $\gamma=0.25$+$k=2$  & 49.14 \\
                   &                                 & SW (ours) & $\gamma=0.25$+$k=5$  & 43.01   & 19.48   & 34.39   &                          &                                 & SW (ours) & $\gamma=0.25$+$k=5$  & 49.60 \\
                   &                                 & SW (ours) & $\gamma=0.25$+$k=10$ & 43.26   & 19.68   & 34.69   &                          &                                 & SW (ours) & $\gamma=0.25$+$k=10$ & 51.29 \\ \cline{3-7} \cline{10-12} 
                   &                                 & OW       & $\gamma==0.5$        & 40.07   & 16.51   & 31.49   &                          &                                 & OW       & $\gamma==0.5$        & 44.58 \\
                   &                                 & SW (ours) & $\gamma=0.5$+$k=1$   & 43.25   & 19.63   & 34.70   &                          &                                 & SW (ours) & $\gamma=0.5$+$k=1$   & 49.66 \\
                   &                                 & SW (ours) & $\gamma=0.5$+$k=2$   & 43.41   & 19.82   & 34.94   &                          &                                 & SW (ours) & $\gamma=0.5$+$k=2$   & 50.77 \\
                   &                                 & SW (ours) & $\gamma=0.5$+$k=5$   & 43.65   & 20.14   & 35.18   &                          &                                 & SW (ours) & $\gamma=0.5$+$k=5$   & 51.50 \\
                   &                                 & SW (ours) & $\gamma=0.5$+$k=10$  & 43.83   & 20.39   & 35.41   &                          &                                 & SW (ours) & $\gamma=0.5$+$k=10$  & 52.50 \\ \cline{3-7} \cline{10-12} 
                   &                                 & OW       & $\gamma==0.75$       & 42.02   & 18.43   & 33.27   &                          &                                 & OW       & $\gamma==0.75$       & 47.31 \\
                   &                                 & SW (ours) & $\gamma=0.75$+$k=1$  & 44.24   & 20.80   & 35.82   &                          &                                 & SW (ours) & $\gamma=0.75$+$k=1$  & 52.59 \\
                   &                                 & SW (ours) & $\gamma=0.75$+$k=2$  & 44.27   & 20.84   & 35.86   &                          &                                 & SW (ours) & $\gamma=0.75$+$k=2$  & 53.27 \\
                   &                                 & SW (ours) & $\gamma=0.75$+$k=5$  & 44.49   & 21.05   & 36.04   &                          &                                 & SW (ours) & $\gamma=0.75$+$k=5$  & 54.04 \\
                   &                                 & SW (ours) & $\gamma=0.75$+$k=10$ & 44.45   & 21.11   & 36.10   &                          &                                 & SW (ours) & $\gamma=0.75$+$k=10$ & 54.45 \\ \cline{2-7} \cline{9-12} 
                   & \multirow{16}{*}{Flan-T5-small} & NW       & -                    & 33.57   & 12.00   & 26.50   &                          & \multirow{16}{*}{Flan-T5-small} & NW       & -                    & 56.41 \\ \cline{3-7} \cline{10-12} 
                   &                                 & OW       & $\gamma==0.25$       & 29.53   & 8.55    & 22.69   &                          &                                 & OW       & $\gamma==0.25$       & 37.93 \\
                   &                                 & SW (ours) & $\gamma=0.25$+$k=1$  & 32.35   & 10.68   & 25.26   &                          &                                 & SW (ours) & $\gamma=0.25$+$k=1$  & 48.45 \\
                   &                                 & SW (ours) & $\gamma=0.25$+$k=2$  & 32.63   & 10.91   & 25.4    &                          &                                 & SW (ours) & $\gamma=0.25$+$k=2$  & 49.47 \\
                   &                                 & SW (ours) & $\gamma=0.25$+$k=5$  & 32.61   & 10.99   & 25.55   &                          &                                 & SW (ours) & $\gamma=0.25$+$k=5$  & 49.73 \\
                   &                                 & SW (ours) & $\gamma=0.25$+$k=10$ & 32.73   & 11.08   & 25.67   &                          &                                 & SW (ours) & $\gamma=0.25$+$k=10$ & 50.03 \\ \cline{3-7} \cline{10-12} 
                   &                                 & OW       & $\gamma==0.5$        & 30.51   & 9.13    & 23.53   &                          &                                 & OW       & $\gamma==0.5$        & 40.88 \\
                   &                                 & SW (ours) & $\gamma=0.5$+$k=1$   & 32.80   & 11.08   & 25.74   &                          &                                 & SW (ours) & $\gamma=0.5$+$k=1$   & 50.05 \\
                   &                                 & SW (ours) & $\gamma=0.5$+$k=2$   & 33.00   & 11.26   & 25.92   &                          &                                 & SW (ours) & $\gamma=0.5$+$k=2$   & 51.30 \\
                   &                                 & SW (ours) & $\gamma=0.5$+$k=5$   & 33.12   & 11.40   & 25.98   &                          &                                 & SW (ours) & $\gamma=0.5$+$k=5$   & 51.65 \\
                   &                                 & SW (ours) & $\gamma=0.5$+$k=10$  & 33.15   & 11.39   & 26.01   &                          &                                 & SW (ours) & $\gamma=0.5$+$k=10$  & 52.16 \\ \cline{3-7} \cline{10-12} 
                   &                                 & OW       & $\gamma==0.75$       & 31.86   & 10.24   & 24.91   &                          &                                 & OW       & $\gamma==0.75$       & 47.19 \\
                   &                                 & SW (ours) & $\gamma=0.75$+$k=1$  & 33.30   & 11.53   & 26.27   &                          &                                 & SW (ours) & $\gamma=0.75$+$k=1$  & 52.84 \\
                   &                                 & SW (ours) & $\gamma=0.75$+$k=2$  & 33.36   & 11.60   & 26.27   &                          &                                 & SW (ours) & $\gamma=0.75$+$k=2$  & 53.85 \\
                   &                                 & SW (ours) & $\gamma=0.75$+$k=5$  & 33.46   & 11.68   & 26.32   &                          &                                 & SW (ours) & $\gamma=0.75$+$k=5$  & 54.23 \\
                   &                                 & SW (ours) & $\gamma=0.75$+$k=10$ & 33.30   & 11.60   & 26.21   &                          &                                 & SW (ours) & $\gamma=0.75$+$k=10$ & 54.29 \\ \cline{2-7} \cline{9-12} 
                   & \multirow{16}{*}{Flan-T5-base}  & NW       & -                    & 39.51   & 16.92   & 31.89   &                          & \multirow{16}{*}{Flan-T5-base}  & NW       & -                    & 59.77 \\ \cline{3-7} \cline{10-12} 
                   &                                 & OW       & $\gamma==0.25$       & 34.47   & 12.00   & 26.94   &                          &                                 & OW       & $\gamma==0.25$       & 41.83 \\
                   &                                 & SW (ours) & $\gamma=0.25$+$k=1$  & 37.66   & 14.92   & 29.91   &                          &                                 & SW (ours) & $\gamma=0.25$+$k=1$  & 51.56 \\
                   &                                 & SW (ours) & $\gamma=0.25$+$k=2$  & 37.85   & 15.06   & 30.12   &                          &                                 & SW (ours) & $\gamma=0.25$+$k=2$  & 51.53 \\
                   &                                 & SW (ours) & $\gamma=0.25$+$k=5$  & 38.23   & 15.40   & 30.55   &                          &                                 & SW (ours) & $\gamma=0.25$+$k=5$  & 52.15 \\
                   &                                 & SW (ours) & $\gamma=0.25$+$k=10$ & 38.24   & 15.45   & 30.53   &                          &                                 & SW (ours) & $\gamma=0.25$+$k=10$ & 51.99 \\ \cline{3-7} \cline{10-12} 
                   &                                 & OW       & $\gamma==0.5$        & 35.23   & 12.57   & 27.52   &                          &                                 & OW       & $\gamma==0.5$        & 45.42 \\
                   &                                 & SW (ours) & $\gamma=0.5$+$k=1$   & 38.34   & 15.45   & 30.66   &                          &                                 & SW (ours) & $\gamma=0.5$+$k=1$   & 51.98 \\
                   &                                 & SW (ours) & $\gamma=0.5$+$k=2$   & 38.49   & 15.72   & 30.82   &                          &                                 & SW (ours) & $\gamma=0.5$+$k=2$   & 52.38 \\
                   &                                 & SW (ours) & $\gamma=0.5$+$k=5$   & 38.79   & 15.91   & 31.02   &                          &                                 & SW (ours) & $\gamma=0.5$+$k=5$   & 52.89 \\
                   &                                 & SW (ours) & $\gamma=0.5$+$k=10$  & 38.67   & 15.89   & 31.01   &                          &                                 & SW (ours) & $\gamma=0.5$+$k=10$  & 53.27 \\ \cline{3-7} \cline{10-12} 
                   &                                 & OW       & $\gamma==0.75$       & 36.98   & 14.13   & 29.34   &                          &                                 & OW       & $\gamma==0.75$       & 50.39 \\
                   &                                 & SW (ours) & $\gamma=0.75$+$k=1$  & 38.92   & 16.15   & 31.26   &                          &                                 & SW (ours) & $\gamma=0.75$+$k=1$  & 55.19 \\
                   &                                 & SW (ours) & $\gamma=0.75$+$k=2$  & 39.00   & 16.31   & 31.38   &                          &                                 & SW (ours) & $\gamma=0.75$+$k=2$  & 55.28 \\
                   &                                 & SW (ours) & $\gamma=0.75$+$k=5$  & 39.13   & 16.35   & 31.47   &                          &                                 & SW (ours) & $\gamma=0.75$+$k=5$  & 55.82 \\
                   &                                 & SW (ours) & $\gamma=0.75$+$k=10$ & 39.03   & 16.34   & 31.44   &                          &                                 & SW (ours) & $\gamma=0.75$+$k=10$ & 55.47 \\ \hline
\end{tabular}}

\caption{Complete results on the XSUM and WebNLG datasets.}
\end{table*}

\end{document}